\documentclass[letterpaper, 10 pt, conference]{ieeeconf}
\IEEEoverridecommandlockouts    
\usepackage{graphics}           
\usepackage{times}              
\usepackage{amsmath}            
\usepackage{amssymb}            
\usepackage{graphicx}
\usepackage{algorithm}
\usepackage[noend]{algpseudocode}
\usepackage{booktabs}
\usepackage{color}
\usepackage{listings}
\usepackage{subfiles}
\usepackage{hyperref}
\usepackage[nocompress]{cite} 
\definecolor{instructioncolor}{rgb}{.5,.5,.5}

\usepackage[font=small]{caption}

\def\secref#1{Sec.~\ref{#1}}
\def\figref#1{Fig.~\ref{#1}}
\def\tabref#1{Tab.~\ref{#1}}
\def\eqref#1{Eq.~(\ref{#1})}

\makeatletter
\usepackage{xspace}
\DeclareRobustCommand\onedot{\futurelet\@let@token\@onedot}
\def\@onedot{\ifx\@let@token.\else.\null\fi\xspace}
 
\def\ie{i.e\onedot}

\def\etal{{et al}\onedot}
\makeatother

\def\etalcite#1{\etal~\cite{#1}}

\usepackage{array}
\newcolumntype{L}[1]{>{\raggedright\let\newline\\\arraybackslash\hspace{0pt}}m{#1}}
\newcolumntype{C}[1]{>{\centering\let\newline\\\arraybackslash\hspace{0pt}}m{#1}}
\newcolumntype{R}[1]{>{\raggedleft\let\newline\\\arraybackslash\hspace{0pt}}m{#1}}

\newcommand{\RR}{\mathbb{R}}

\newcommand{\abs}[1]{|#1|}

\renewcommand{\b}[1]{\mbox{\boldmath$#1$}}

\renewcommand{\v}[1]{{\b #1}} 

\newcommand{\by}{\b y}

\usepackage{multirow}
\usepackage{hyperref}

\title{\LARGE \bf Improving Indoor Localization Accuracy\\by Using an Efficient Implicit Neural Map Representation}

\author{Haofei Kuang \quad Yue Pan \quad Xingguang Zhong \quad Louis Wiesmann \quad Jens Behley \quad Cyrill Stachniss
  \thanks{All authors are with the Center for Robotics, University of Bonn, Germany. Cyrill Stachniss is additionally with the Department of Engineering Science at the University of Oxford, UK, and with the Lamarr Institute for Machine Learning and Artificial Intelligence, Germany.}%
  \thanks{This work has partially been funded 
  by the Deutsche Forschungsgemeinschaft (DFG, German Research Foundation) under Germany's Excellence Strategy, EXC-2070 -- 390732324 -- PhenoRob,
  and by the German Federal Ministry of Education and Research~(BMBF) in the project ``Robotics Institute Germany'' under grant No.~16ME0999.
  }%
}

\begin{document}
\thispagestyle{empty}
\pagestyle{empty}
\maketitle

\begin{abstract}
  %
  Globally localizing a mobile robot in a known map is often a foundation for enabling robots to navigate and operate autonomously.
  In indoor environments, traditional Monte Carlo localization based on occupancy grid maps is considered the gold standard, but its accuracy is limited by the representation capabilities of the occupancy grid map.
  In this paper, we address the problem of building an effective map representation that allows to accurately perform probabilistic global localization.
  To this end, we propose an implicit neural map representation that is able to capture positional and directional geometric features from 2D LiDAR scans to efficiently represent the environment and learn a neural network that is able to predict both, the non-projective signed distance and a direction-aware projective distance for an arbitrary point in the mapped environment.
  This combination of neural map representation with a light-weight neural network allows us to design an efficient observation model within a conventional Monte Carlo localization framework for pose estimation of a robot in real time.
  We evaluated our approach to indoor localization on a publicly available dataset for global localization and the experimental results indicate that our approach is able to more accurately localize a mobile robot than other localization approaches employing occupancy or existing neural map representations.
  In contrast to other approaches employing an implicit neural map representation for 2D LiDAR localization, our approach allows to perform real-time pose tracking after convergence and near real-time global localization.
  The code of our approach is available at: \url{https://github.com/PRBonn/enm-mcl}.
\end{abstract}

\section{Introduction}
\label{sec:intro}

Estimating the state of a robot in terms of its position and orientation in an environment is crucial to enable robots to operate autonomously, but it is also a key requirement for realizing autonomous navigation and planning.
Localization in a pre-built map is a common way to realize state estimation for robotic systems, where probabilistic methods using various map representations are often employed.
In indoor environments, localization via external sources of global positioning information, such as GNSS data, is typically not available, therefore, global localization must rely on onboard sensors, like wheel odometry and 2D LiDAR sensors. 
A gold standard approach for that is Monte Carlo localization~\cite{dellaert1999icra}.

In this paper, we investigate the problem of building an effective map for achieving accurate and efficient global localization and ego-pose tracking in indoor environments.
Specifically, we aim to build a map representation that can be used for implementing an accurate and computationally efficient probabilistic localization approach, where we employ a conventional Monte Carlo localization~(MCL) and use our map representation for an efficient observation model.

A classical approach to build an effective map representation is occupancy grid mapping~\cite{thrun2005probrobbook} often used to realize MCL~\cite{fox1999aaai, fox2001neurips, zimmerman2022iros} in indoor environments.
Recently, several approaches~\cite{kuang2023ral,wiesmann2023ral} investigated the usage of novel implicit neural map representations to replace the commonly employed occupancy grid maps due to promising prospects regarding representing scene details continuously with a comparably small memory footprint and scene completion capabilities~\cite{zhong2023icra}.
While these approaches demonstrate the advantages of an implicit map representation over conventional occupancy maps, they typically require significant computational resources for training and deployment in global localization, especially in large-scale environments.

\begin{figure}[t]
	\centering
	\includegraphics[width=0.92\linewidth]{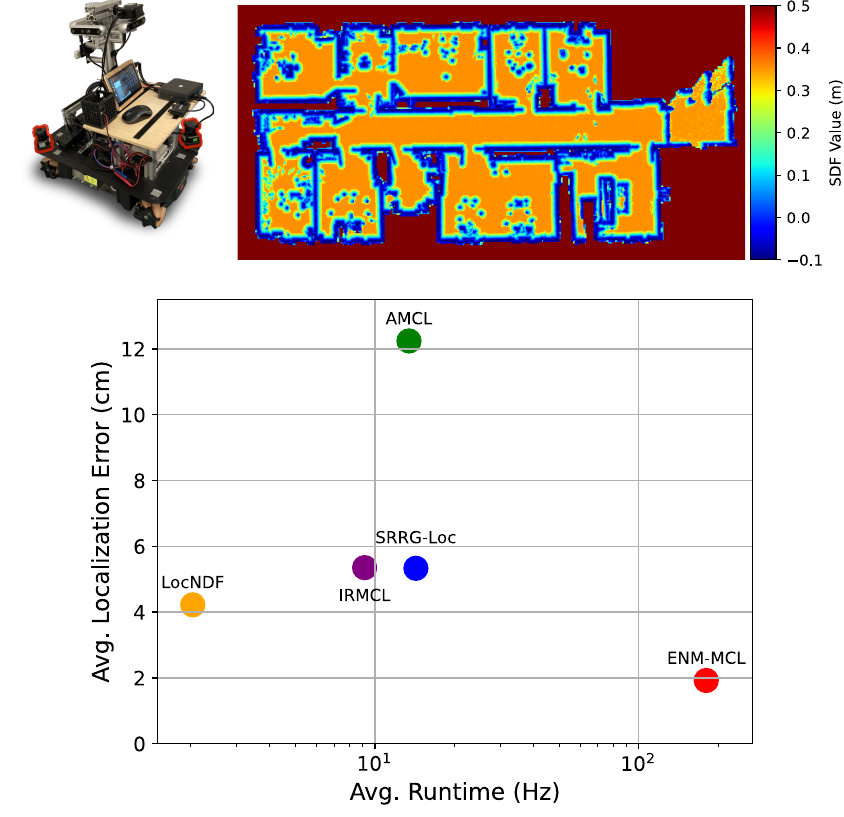}
	\caption{We learn a neural network to represent the surfaces of environments. Our method achieves both efficiency and accuracy for indoor localization by integrating our proposed efficient neural map representation into a Monte Carlo localization system. We show the average runtime for each method in sequence 1 of the in-house dataset.}
	\label{fig:motivation}
	\vspace{-0.6cm}
\end{figure}


To address these key limitations of current implicit neural map representation-based global localization methods~\cite{kuang2023ral,wiesmann2023ral}, we propose a novel approach that improves the efficiency of using the map representation in an MCL-based localization, while simultaneously increasing the localization accuracy at the same time, as we show in~\figref{fig:motivation}.

The main contribution of this paper is a novel implicit map representation, which we call efficient neural map~(ENM), that allows to learn the surface information of a mapped environment through a dense feature grid combined with a light-weight neural network enabling efficient global localization in large indoor environments with a large number of particles. 
The light-weight neural network learns an approximate regular (non-projective) signed distance field~(SDF), but is also able by incorporating information of the ray direction to accurately predict a direction-aware projective SDF~(PSDF) capturing both, geometric and directional features of the environment.
We integrate our ENM representation into the existing MCL framework by replacing the occupancy grid map used to compute a beam end point-based observation model~\cite{thrun2005probrobbook}, resulting in the approach we call ENM-MCL.
Our experimental evaluation on multiple publicly available sequences for indoor localization indicates that our ENM-MCL is able to more accurately localize a mobile robot than other probabilistic localization approaches employing occupancy or so far used neural representations.
Furthermore, the results show that ENM-MCL is able to perform real-time pose tracking after convergence, while allowing near real-time global localization.

In sum, we make three key claims:
(i) Our map representation, ENM, used by ENM-MCL achieves more accurate global localization compared to other localization approaches employing conventional occupancy grid maps or neural map representations;
(ii) Our ENM-MCL system operates at real time for pose tracking but also enables efficient global localization even with a large number of particles;
(iii) Our ENM-MCL converges quickly to the correct pose estimate enabling fast and reliable localization in indoor environments.
We will publish the source code of our ENM-MCL.

\section{Related Work}
\label{sec:related}


Pose tracking and global localization for mobile robots are key research areas in robotics as determining the location of a robot in a given map is crucial for many downstream tasks.
Regarding global localization for mobile robots, Monte Carlo localization is a gold standard method, which exploits a particle filter to estimate the robot's state in a probabilistic manner and has been implemented with various sensors, such as 2D LiDARs~\cite{dellaert1999icra, fox2001neurips, stachniss2005aaai} and cameras~\cite{bennewitz2006euros, zimmerman2023ral}.
To improve the efficiency of the MCL algorithm, Fox~\etalcite{fox2001neurips} propose an adaptive sampling strategy, which adapt the size of the sample set on the fly.
Traditionally, MCL methods for LiDAR-based localization often rely on occupancy grid maps~\cite{fox2001neurips, zimmerman2022iros, chen2021icra} for 2D LiDARs, but also for 3D LiDARs~\cite{lu2019cvpr, chen2020iros}.

The representational capacity of occupancy maps is constrained by their pre-defined grid resolution, where simply increasing the resolution can substantially increase memory consumption.
To overcome this limitation of a fixed resolution, some approaches for LiDAR-based localization have been proposed to construct a multi-resolution map~\cite{droeschel2018icra} or learn continuous implicit map representations via Gaussian processes~\cite{ocallaghan2012ijrr, yuan2018icarcv}, reproducing kernel Hilbert maps~\cite{ramos2016ijrr}.
Nonetheless, these methods are quite time-consuming and challenging to directly apply to real-world scenarios.

Recently, implicit neural map representations have gained popularity due to their compactness and continuous representational capacity. 
These representations have been used to model complex geometric information, such as learning a radiance field~\cite{mildenhall2020eccv, rematas2022cvpr, barron2022cvpr} from images to representing the 3D world~\cite{park2019cvpr, ortiz2022rss}.
Consequently, NeRFs have been employed for visual localization tasks, such as global localization~\cite{maggio2023icra} or re-localization~\cite{liu2023icra}.
Other methods~\cite{park2019cvpr, ortiz2022rss} learn a signed distance field from LiDAR or RGB-D camera as a more accurate geometric representation of the environment. 
Moreover, some works~\cite{zhi2021iccv, yuan2024tro} use the neural network not only to encode the geometry or appearance of the scene, but also the semantics of the environment.
Due to their high representational capacity of the employed neural network, several works~\cite{kuang2023ral, wiesmann2023ral} use a single MLP to represent the entire scene and integrate the implicit neural map representation into an observation model of MCL to achieve global localization.
Although these methods exhibit good map representation capabilities, the reconstruction process is very time-consuming and the time needed to querying the representation does not meet the required efficiency for real-time localization.

To address the challenge with respect to efficiency, recent works~\cite{zhong2023icra, zhu2022cvpr, pan2024tro} have introduced feature fields combined with shallow MLPs as a representation of the environment. 
This approach not only accelerates convergence of training process but also enhances the quality and capacity of the map representation. 
However, these methods have primarily been applied to 3D reconstruction tasks and have not yet been efficiently deployed in 2D indoor localization scenarios.

In this work, we leverage the high capacity of neural networks to learn both the non-projective SDF and a direction-aware projective SDF by encoding the ray direction of 2D LiDAR into the neural map representation. 
Additionally, we propose a novel implicit representation based on a dense feature grid combined with a light-weight neural network, which substantially reduces computational cost, especially for querying the neural map representation. 
Our novel implicit neural map representation enables more detailed modeling of the environment while improving the accuracy and efficiency of global localization using 2D LiDAR scans.

\begin{figure*}[t]
	\centering
	\includegraphics[width=1.0\linewidth]{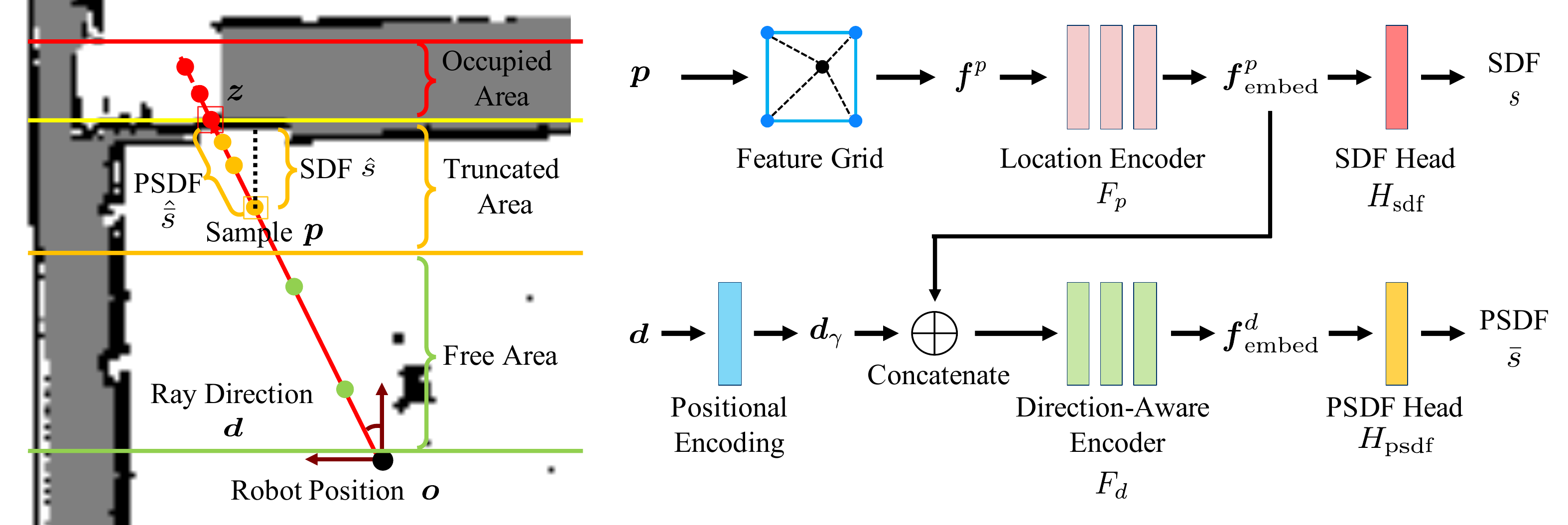}
	\caption{Overview of our approach for jointly predicting the non-projective SDF and direction-aware projective SDF values using our proposed efficient neural map representation. We sample several positions along a LiDAR ray and input the 2D position $\v{p} = (x, y)^{\top} $ as well as the corresponding ray direction $ \v{d} = (d_{x}, d_{y})^{\top} $ into our ENM model to estimate the SDF and PSDF. The predictions of the SDF and PSDF values by the neural network are supervised with the ground truth SDF/PSDF values from 2D LiDAR measurements. 
	}
	\label{fig:overview}
	\vspace{-0.6cm}
\end{figure*}

\section{Our Approach for Global Localization}
\label{sec:main}


The goal of our approach is to accurately localize a mobile robot in a given map by learning an efficient neural map representation of the environment.
More specifically, we propose to represent the environment via a feature grid-based representation, which is learned from 2D LiDAR scans recorded in a mapping session and allows to predict the signed distance to the surface at an arbitrary position via a shallow multi-layer perceptron~(MLP), see~\secref{sec:main_enm}.
For learning the implicit neural map representation, we supervise learnable features of the feature grid and the weights of the employed MLP with real measurements from a 2D LiDAR as presented in~\secref{sec:main_opt}.
After that, we use the efficient neural map representation in an MCL-based localization approach, where we use the estimated SDF for an observation model to update the weights of the particles, as described in more detail in~\secref{sec:main_mcl}.

\subsection{Efficient Neural Map Representation}
\label{sec:main_enm}

For our map representation, we want to leverage a neural representation that allows for estimating the signed distance to the surface in the environment.
More specifically, we propose an implicit neural surface representation, which is a function $F_{\Theta, \mathbf{G}}$ that takes the 2D location vector $\v{p} = (x, y)^{\top}$ and 2D Cartesian unit vector $\v{d} = (d_{x}, d_{y})^{\top}$ as inputs, and predicts the corresponding non-projective signed distance field (SDF) value~$s$  and direction-aware projective signed distance field (PSDF) value~$ \bar{s} $. 
The SDF value represents the shortest distance from the point $\v{p}$ to the nearest surface in the environment and the PSDF is the distance from a point $\v{p}$ to the surface along the specific ray direction $\v{d}$. 
Formally, we have the following:
\begin{equation}
	s, \bar{s} = F_{\Theta, \mathbf{G}}(\v{p}, \v d).
\end{equation}

We represent $F_{\Theta, \mathbf{G}}$ by a MLP, where $\Theta$ represents the weights of the MLP and $\mathbf{G}$ corresponds to a dense grid of learnable feature vectors.
By combining a light-weight network with a dense feature grid, we can substantially reduce the computation-cost meanwhile maintaining the quality of the map model, which supports to use it with a particle filter for localization in real time.

Our dense feature grid $\mathbf{G}$ has a given grid resolution~$\Delta_\mathbf{G}$. 
Each grid corner at the location coordinate $(i, j)$ stores a~\mbox{$D$-dimensional} feature vector $\v{c}_{i,j}^{g} \in \mathbb{R}^{D}$.
During operation, the input 2D location $\v{p}$ is first encoded into a location feature $\v{f}^{p} \in \mathbb{R}^{D}$ by performing bilinear interpolation on the dense feature grid~$\mathbf{G}$.
We use the neural network to estimate the SDF $s$ and PSDF $\overline{s}$ of input 2D location $\v{p}$ and ray direction $\v{d}$, which consist of two branches, as shown in~\figref{fig:overview}.

In the SDF branch, the location feature $\v{f}^{p}$ is first processed by an encoder $F_p$, a 3-layer MLP, to extract the positional embedding~$\v{f}_{\text{embed}}^{p}$.
Then, $\v{f}_{\text{embed}}^{p}$ is decoded into the $s$ value of the input location by the SDF head $H_{\text{sdf}}$, which is a 1-layer MLP.
Formally, we have the following:
\begin{align}
	\v{f}_{\text{embed}}^{p} &= F_{p}(\v{f}^{p}), \\
	s &= H_{\text{sdf}}(\v{f}_{\text{embed}}^{p}).
\end{align}


In the PSDF branch, the extracted positional embedding~$\v{f}_{\text{embed}}^{p}$ is concatenated with the positional encoding of the directional vector $\v{d}_{\gamma}$ and fed into another 3-layer MLPs $F_{d}$, to extract the direction-aware embedding $\v{f}_{\text{embed}}^{d}$, and then predict the $\bar{s}$ value of the input location and direction by a PSDF head $H_{\text{psdf}}$, which is also a 1-layer MLP, as follows:
\begin{align}
	\v{f}_{\text{embed}}^{d} &= F_{d}(\v{f}_{\text{embed}}^{p} \oplus \v{d}_{\gamma}), \\
	\bar{s} &= H_{\text{psdf}}(\v{f}_{\text{embed}}^{d}),
\end{align}
where $\oplus$ is the concatenation of two vectors.
To enable the model to capture high-frequency geometric features of the ray direction, we encode the directional vector $\v d$ via a positional encoding $\gamma$, where we apply $\gamma$ to each component~\cite{mildenhall2020eccv}, \ie, $\v{d}_{\gamma} = (\gamma(d_{x}), \gamma(d_{y}))^{\top}$. 
The positional encoding function $\gamma: \RR \mapsto \RR^{2L+1}$ is defined as:
\begin{align}
	\gamma(d) &= \left(d, \sin(2^{0}d), \cos(2^{0}d), \dots, \sin(2^{L-1}d), \cos(2^{L-1}d)\right),
\end{align}
where $L$ is the number of frequency bands used. 

The networks $F_{p}$ and $F_{d}$ are shallow MLPs. 
Each hidden layer has $D$ neurons and $D + 2(2L + 1)$ neurons for $F_{p}$ and $F_{d}$, respectively.
$D$ is the dimension of the feature vector from $\mathbf{G}$ and $L$ is the bandwidth of $\gamma$.
In our model, we set $D = 4$ and $L = 4$, and each layer is followed by a ReLU activation. 
These shallow MLPs enable our model to run in real-time, even with a large number of particles.

\subsection{Learning the ENM from 2D LiDAR Data}
\label{sec:main_opt}

We learn our ENM representation using 2D LiDAR data from scans recorded in a mapping run. 
Given a posed 2D LiDAR scan with  a set of rays $\mathcal{B} = \{(r_j, \v d_j)\}$, where $\v{d}_j$ are ray directions and $r_j$ are range readings, we sample positions on these rays for a training set $\mathcal{S}$ for each scan.


We sample training data by selecting points along each LiDAR ray within three regions: the truncated space, occupied space, and free space, see~\figref{fig:overview} left. 
The truncated space is the area in front of the surface, and the occupied space is the area behind the surface.
For each ray, we randomly sample $M_t$ samples in the truncated space, denoted as $\mathcal{S}_t$, and fewer $M_o$ samples in the occupied space, denoted as $\mathcal{S}_o$, to encourage the model to learn the surface features.
Furthermore, we sample a small number of $M_s$ samples in the free space, denoted as $\mathcal{S}_f$, as these areas contribute less to learn the fine geometric details. 
The set of sampled points along each ray of a single LiDAR scan is then given by $\mathcal{S} = \mathcal{S}_{t} \cup \mathcal{S}_{o} \cup \mathcal{S}_{f}$.
\figref{fig:overview} shows the sampling process visually.

To optimize the weights and learnable feature vectors of~$F_{\Theta, \mathbf{G}}$, we supervise the predicted~$s$ and $\bar{s}$ values using the generated ground truth from the training samples.  
Regarding the PSDF loss $\mathcal{L}_{\text{psdf}}$, we can directly generate the ground-truth projective SDF values from the range readings of each LiDAR ray.  
Let $\v{p}$ be a sample on the LiDAR ray, then the ground-truth PSDF value $\hat{\bar{s}}$ is given by the distance between real measured distance $r$ and the distance between $\v{p}$  and the LiDAR origin $\v{o}$:
\begin{equation}
	\hat{\bar{s}} = r - \| \v{p} - \v{o} \|_2.
\end{equation}

The loss $\mathcal{L}_{\text{psdf}}$ for the projected SDF is then computed using only the sampled points near the surface, $\mathcal{S}_t \cup \mathcal{S}_o$, as:
\begin{equation}
	\mathcal{L}_{\text{psdf}} = \frac{1}{|\mathcal{S}_{t}| + |\mathcal{S}_{o}|} \sum_{\left(\v{p}, \v{d}\right) \in \mathcal{S}_{t} \cup \mathcal{S}_{o}} | \bar{s} - \hat{\bar{s}} |,
\end{equation}
where $\v{d}$ is the corresponding ray direction of sample $\v{p}$ and~$\bar{s}$ is the predicted PSDF from $F_{\Theta, \mathbf{G}}(\v{p}, \v{d})$.


Regarding the SDF loss $\mathcal{L}_{\text{sdf}}$, we activate the predicted SDF values using a sigmoid function~$\sigma$, and minimize a binary cross-entropy loss as the SDF objective $\mathcal{L}_{\text{sdf}}$, formulated as:
\begin{align}
	\mathcal{L}_{\text{sdf}} = - \frac{1}{|\mathcal{S}|} \sum_{(\v{p}, \v{d}) \in \mathcal{S}} \Big( &\sigma(\hat{s}) \log \sigma(s) \nonumber \\
	&+ (1 - \sigma(\hat{s})) \log(1 - \sigma(s)) \Big),
\end{align}
where \mbox{$s = F_{\Theta, \mathbf{G}}(\v{p}, \v{d})$} is the predicted signed distance value for the input sample, and $\sigma(\hat{s}) \in [0, 1]$ is the ground-truth value indicating the probability of the point being inside or outside the surface.
Since we cannot directly obtain the non-projective SDF value from LiDAR range readings, we supervise the SDF prediction in an approximate manner, \ie, $ \sigma(\hat{s}) \approx \sigma(\hat{\bar{s}}) $, inspired by prior work~\cite{zhong2023icra}.

Additionally, we incorporate an Eikonal loss $\mathcal{L}_{\text{eikonal}}$ to regularize the SDF~\cite{zhong2023icra, pan2024tro}. 
The Eikonal loss ensures that the gradient of the predicted SDF $\nabla s$ satisfies the Eikonal equation $\| \nabla s \| = 1$, which helps to enforce SDF smoothness and prevents the SDF from unrealistic deformations, leading to a more accurate neural map representation.
The Eikonal loss $\mathcal{L}_{\text{eikonal}}$ is defined as:
\begin{equation}
	\mathcal{L}_{\text{eikonal}} = \frac{1}{|\mathcal{S}_{t}| + |\mathcal{S}_{o}|} \sum_{\left(\v{p}, \v{d}\right) \in \mathcal{S}_{t} \cup \mathcal{S}_{o}} \left( \| \nabla s_i \| - 1 \right)^2.
\end{equation}

By minimizing this loss, we ensure that the predicted SDF adheres to the properties of a signed distance function, improving both stability and accuracy in the learned map representation.
To sum up, our final loss objective is:
\begin{equation}
	\mathcal{L}_{\text{final}} = \mathcal{L}_{\text{sdf}} + \mathcal{L}_{\text{psdf}} + \beta \mathcal{L}_{\text{eikonal}}.
\end{equation}

We optimize the loss function of the ENM model using the Adam optimizer~\cite{kingma2015iclr} with a learning rate of $0.001$, and we use $\beta = 0.1$ for the Eikonal loss.
$\mathcal{S}$. 
We sample $M_{t} = 6$ positions in the truncated space, $M_{o} = 4$ positions in the occupied space, and $M_{f} = 5$ positions in the free space.
The training process runs for 5,000 iterations, at each iteration we randomly sample batch of rays from all training LiDAR scans with a batch size of 2,048. 

\begin{figure}[t]
	\centering
	\includegraphics[width=1.0\linewidth]{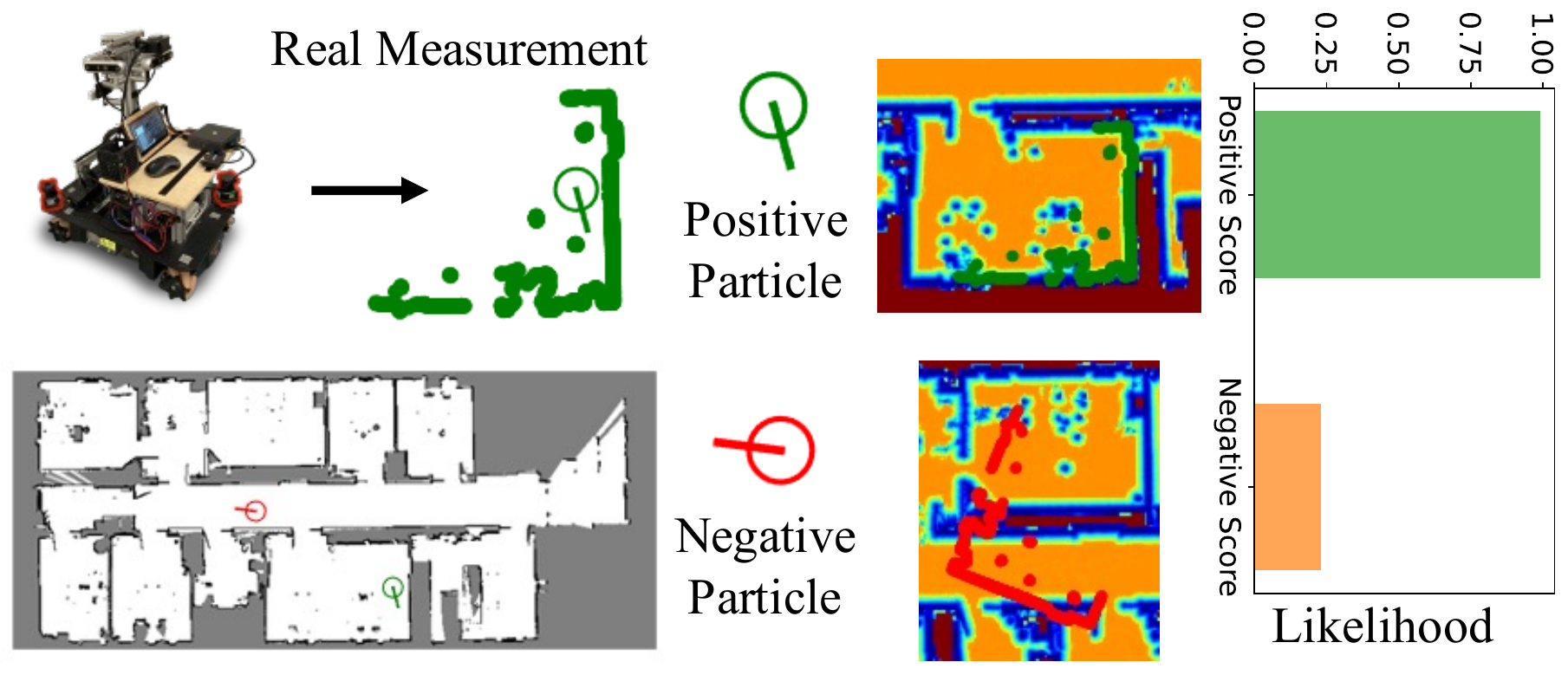}
	\vspace{-0.4cm}
	\caption{The observation model based on the ENM representation. We estimate the SDF and PSDF values for all beam end-points of the particles using the ENM model, and update the particles' weights by computing the likelihood based on the SDF and PSDF values of the beam end-points. The positive particle has higher likelihood which scans are aligned with the zero level of distance field. 
	}
	\label{fig:obs_model}
	\vspace{-0.4cm}
\end{figure}

\subsection{Efficient Neural Map-based MCL}
\label{sec:main_mcl}

Given our new map representation, we need to integrate it into MCL as our localization approach, ENM-MCL.
We want to estimate the SE(2) state $\v{x}_{t} = (x, y, \theta)_{t}^{\top}$ of the robot at time~$t$ defined by the 2D position $(x, y)^{\top}$ and the orientation $\theta \in [-\pi, \pi]$. 
To this end, we estimate the posterior $p(\v x_t| \v z_t, m)$ of the robot state $\v x_t$ at time $t$ with observations $\v z_t$ and map $m$ using a recursive Bayes filter~\cite{thrun2005probrobbook}, which is realized in MCL~\cite{dellaert1999icra} via a particle filter.
The particle filter approximates $p(\v x_t| \v z_t, m)$ by a set of particles $\mathcal{N} = \{(\v{x}_{t}^{n}, w_{t}^{n})\}$, $|\mathcal{N}| = N$, where each particle is a hypothesis of the robot's state $\v{x}_{t}^{n} = (x_{t}^{n}, y_{t}^{n}, \theta_{t}^{n})^{\top}$ with its corresponding weight $w_{t}^{n}$. 
The weights $w_{t}^{n}$ are updated based on the likelihood $p(\v z_t|\v x^n_t, m)$ of the observation $\v z_t$, commonly called observation model.


We use as observation model the conventional beam end-point model of MCL, but adapt it to employ our map representation.
Here, we want to achieve that the observation model results in a higher likelihood if a particle hypothesis is closer to the real robot state, which means that the beam end points have the small SDF and PSDF values with our map representation.
Specifically, we transfer the measured LiDAR scan to a particle's state $\v{x}_{t}^{n}$ at time~$t$ and then check the SDF and PSDF value of end-point $\v{z}_{t}^{i}$ for each ray $(r_{t}^{i}, \v{d}_{t}^{i}) \in \mathcal{B}^{t}$, formally as:
\begin{equation}
	\v{z}_{t}^{i} = \mathbf{R}_{t}^{n} (r_{t}^{i} \v{d}_{t}^{i}) + t_{t}^{n},
\end{equation}
where $\mathbf{R}_{t}^{n} \in \RR^{2\times 2}$ is the 2D rotation matrix for angle $\theta_{t}^{n}$, and \mbox{$t_{t}^{n} = (x_{t}^{n}, y_{t}^{n})^{\top}$}.
Then, we exploit our ENM model $F_{\Theta, \mathbf{G}}$ to estimate the SDF value $s_{t}^{i}$ of $\v{z}_{t}^{i}$, and the PSDF value $\bar{s}_{t}^{i}$ of $(\v{z}_{t}^{i}, \mathbf{R}^n_t\v d^i_t)$ as:
\begin{equation}
	s_{t}^{i}, \bar{s}_{t}^{i} = F_{\Theta, \mathbf{G}} (\v{z}_{t}^{i}, \mathbf{R}^n_t\v d^i_t),
\end{equation}
where $\mathbf{R}^n_t\v d^i_t$ is $i$-th the ray direction in the reference frame of the particle.
Then, the likelihood of $\v{z}_{t}^{i}$ is given by:
\begin{equation}
	p(\v{z}_{t} \mid \v{x}_{t}^{n}, F_{\Theta, \mathbf{G}}) \propto \exp\left(-\lambda \frac{1}{|\mathcal{B}^{t}|}\sum_{i=1}^{|\mathcal{B}^{t}|} \frac{\abs{s_{t}^{i}} + \abs{\bar{s}_{t}^{i}}}{2}\right).
	\label{eq:obs_model}
\end{equation}

Similar to other localization systems~\cite{grisetti2018github}, we compute an average alignment for each scan to the map.
Furthermore, we average the predictions from the ENM model since it reduces the impact of noise or inaccuracies in either $s_{t}^{i}$ or $\bar{s}_{t}^{i}$, resulting in a more robust and consistent likelihood estimation.


\section{Experimental Evaluation}
\label{sec:exp}

%
The focus of this work is an efficient MCL system using an efficient and accurate neural map representation.
%
We present our experiments to show the capabilities of our method. 
The results of our experiments support our key claims, which are:
(i) the method achieves high accuracy in localization by representing the environment with the ENM model;
(ii) our ENM-MCL system operates in real-time at pose tracking but also enables efficient global localization even with large numbers of particles; 
(iii) our algorithm converges quickly to the correct pose estimation to support rapid and reliable localization.

\begin{table}[t]
	\centering
	\renewcommand\arraystretch{0.95}
		\begin{tabular}{c|c|c|c|c}
			\toprule
			\multirow{2}{*}{Seq}               & \multirow{2}{*}{Method} 		& Location & Yaw & \multirow{2}{*}{\shortstack{Success\\Rate (\%)}} \\
			& 		& RMSE (cm) $\downarrow$ & RMSE (degree) $\downarrow$ & \\
			\midrule
			\multirow{6}{*}{1} 
			& AMCL 		& 12.24 $\pm$ 0.33  & 2.08 $\pm$ 0.08  & 40.0 \\
			& SRRG-Loc 	& 5.33 $\pm$ 0.02 & 0.75 $\pm$ 0.00 & 100.0 \\
			& IRMCL 	 & 5.35 $\pm$ 0.07 & 1.06 $\pm$ 0.01 & 100.0 \\
			& LocNDF 	 & 4.06 $\pm$ 0.06 & 0.60 $\pm$ 0.00 & 60.0 \\
			& ENM-MCL 	 & \textbf{1.96 $\pm$ 0.07}  & \textbf{0.41 $\pm$ 0.00}  & \textbf{100.0} \\
			\midrule
			\multirow{6}{*}{2} 
			& AMCL 		& 10.28 $\pm$ 0.00  & 0.86 $\pm$ 0.00  & 100.0 \\
			& SRRG-Loc 	& 6.59 $\pm$ 0.02 & 1.09 $\pm$ 0.00 & 100.0 \\
			& IRMCL 	 & 5.53 $\pm$ 0.06 & 0.82 $\pm$ 0.01 & 60.0 \\
			& LocNDF 	 & \textbf{4.06 $\pm$ 0.93} & 0.62 $\pm$ 0.03 & 80.0 \\
			& ENM-MCL 	 & 4.18 $\pm$ 0.06  & \textbf{0.60 $\pm$ 0.01}  & \textbf{100.0} \\
			\midrule
			\multirow{6}{*}{3} 
			& AMCL 		& -                 & -                & 0.0 \\
			& SRRG-Loc 	& - & - & 0.0 \\
			& IRMCL 	 & 4.70 $\pm$ 0.13 & 0.77 $\pm$ 0.04 & 100.0 \\
			& LocNDF 	 & 3.85 $\pm$ 0.13 & \textbf{0.72 $\pm$ 0.01} & 100.0 \\
			& ENM-MCL 	 & \textbf{3.05 $\pm$ 0.03}  & 0.74 $\pm$ 0.01  & \textbf{100.0} \\
			\midrule
			\multirow{6}{*}{4} 
			& AMCL 		& -                 & -  & 0.0 \\
			& SRRG-Loc 	& - & - & 0.0 \\
			& IRMCL 	 & 11.72 $\pm$ 0.31 & 1.57 $\pm$ 0.12 & 100.0 \\
			& LocNDF 	 & 10.89 $\pm$ 0.55 & 1.55 $\pm$ 0.16 & 60.0 \\
			& ENM-MCL 	 & \textbf{5.82 $\pm$ 0.12}  & \textbf{0.97 $\pm$ 0.08}  & \textbf{100.0} \\
			\midrule
			\multirow{6}{*}{5} 
			& AMCL 		& -                 & -                & 0.0 \\
			& SRRG-Loc 	& 6.29 $\pm$ 0.06 & 1.03 $\pm$ 0.05 & 100.0 \\
			& IRMCL 	 & 6.12 $\pm$ 0.03 & 1.26 $\pm$ 0.01 & 100.0 \\
			& LocNDF 	 & -                 & -                & 0.0 \\
			& ENM-MCL 	 & \textbf{2.40 $\pm$ 0.02}  & \textbf{0.54 $\pm$ 0.01}  & \textbf{100.0} \\
			\bottomrule           
		\end{tabular}
	\caption{Quantitative results of global localization on the in-house dataset. We report the ATE for both location and orientation RMSE, along with the success rate of each method over five runs. The ATE is only reported if at least one run was successful; otherwise, `-' indicates failure.}
	\label{tab:ipblab_loc}
	\vspace{-0.5cm}
\end{table}

\subsection{Experimental Setup}
\label{sec:exp_setup}

We evaluate our global localization results using the in-house dataset, previously used in prior work~\cite{kuang2023ral, wiesmann2023ral}. 
It was collected with a KuKa YouBot equipped with a UTM-30LX 2D LiDAR and an upward-facing camera (not used for localization). 
The ground-truth robot poses were generated by localizing a large number of AprilTags on the ceiling detected by the upward-facing camera, as described in our prior work~\cite{kuang2023ral}, providing a reference trajectory as near ground truth with a global position error around 1\,cm, often even better. 
The dataset covers different indoor scenes such as offices, a kitchen, and a long corridor. 
It contains a 31,608-frame mapping sequence for training ENM and five test sequences averaging 1,419 frames for localization evaluation.

To demonstrate the localization accuracy of our method, we compare with four existing LiDAR-based localization algorithms: AMCL~\cite{fox2001neurips}, SRRG-Loc~\cite{grisetti2018github}, IRMCL~\cite{kuang2023ral}, and LocNDF~\cite{wiesmann2023ral}.
AMCL and SRRG-Loc are both occupancy grid map-based methods. 
The AMCL is a widely used global localization method from the standard ROS1 implementation.
The SRRG-Loc~\cite{grisetti2018github} is developed by the Sapienza Robust Robotics Group (SRRG), which is a well-designed MCL implementation by Giorgio Grisetti and is based on occupancy grid maps.
In contrast, IRMCL and LocNDF are two state-of-the-art methods based on implicit neural representations. 

\subsection{Global Localization Performance}
\label{sec:exp_results}

The first experiment evaluates the performance of our approach and its outcomes
support the claim that our method achieves the state-of-the-art accuracy for global localization.


Regarding the parameters of our MCL system, we use a large particle number $N =$ 80,000 during the initialization phase, where particles are uniformly distributed across the map with random orientations for global localization, and reduce it to $N =$ 1,000 for efficient pose tracking after convergence.
This adaptive adjustment for the particle number balances the trade-off between reliability and computational cost for the MCL system.
To evaluate the localization results, we compute the absolute trajectory error~(ATE) between the predicted and ground-truth trajectories. 
It includes the RMSE of translation~(in centimeters) and orientation~(in degrees).
To reduce the sensitivity to randomness in the particle filter, we run five times with different random seeds, and take the average as the final ATE for each sequence.

\begin{figure}[t]
	\centering
	\includegraphics[width=1.0\linewidth]{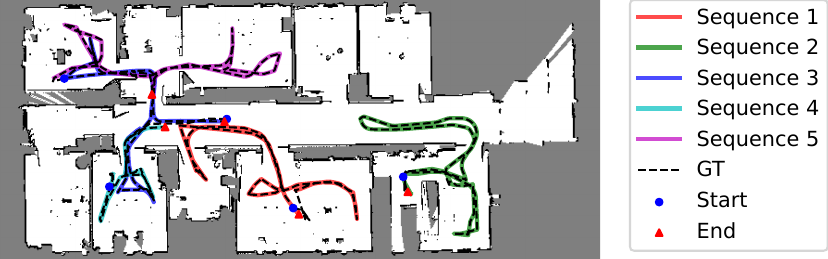}
	\caption{Qualitative global localization results of ENM-MCL on all five sequences of the in-house dataset. We show the predicted trajectories after convergence. The predicted trajectories are mostly aligned with the ground truth, indicating the high accuracy and reliability of our method.}
	\label{fig:trajectory}
	\vspace{-0.5cm}
\end{figure}

The comparison results are shown in \tabref{tab:ipblab_loc}. 
Our method demonstrates better accuracy compared to the baseline methods across all sequences. 
Even in challenging scenarios, such as sequence 3 where the robot starts in a corridor, our method substantially outperforms other approaches, leading to a 35.7\% improvement over IRMCL.
Meanwhile occupancy grid map-based methods failed entirely without parameter tuning in this sequence.
Furthermore, the results highlight that our method is more robust than the baselines, achieving a 100\% success rate on all five sequences. 
This indicates that our model offers a better geometric representation to support more reliable localization in complex indoor environments.
The predicted trajectories of our method are shown in \figref{fig:trajectory}.

\subsection{Runtime Analysis}
\label{sec:exp_runtime}

The second experiment evaluates the runtime for both training and localization, illustrating the efficiency of our approach. 
Specifically, we measure the runtime during performing MCL in initialization and pose tracking phase, and also propose the average frame rate on sequence 1 of the in-house dataset. 
We compare our method with IRMCL~\cite{kuang2023ral} and LocNDF~\cite{wiesmann2023ral}, two other implicit representation-based MCL methods. 
Since all these methods are based on neural networks to build implicit representations, our results support the claim that the ENM architecture can reduce computation costs for real-time applications.
We test all approaches on a desktop computer with a 3.7 GHz CPU and 64 GB memory, and a NVIDIA Quadro RTX 5000 GPU with 16 GB memory.

As shown in~\tabref{tab:speed}, the runtime performance of the tested methods demonstrates the efficiency of our approach. 
We keep the default settings of the number of particle to each method for fair comparison.
Specifically, our ENM-MCL outperforms the other methods in both initialization and pose tracking phases on the dataset. 
During pose tracking, ENM-MCL achieves 180.2 fps using 1,000 particles, making it suitable for real-time operation. 
Although our method requires more particles during the initialization phase (80,000 particles), it still reaches a speed of 4 Hz, showcasing its efficiency even with larger particle sets.


\subsection{Convergence Analysis}
\label{sec:exp_success}

Finally, we analyze our method with respect to its ability for rapid and reliable global localization. 
For fair comparison, we compare the convergence speed of localization with baseline methods in sequence 1 of the dataset, as it is a relatively simple scenario where all baselines succeed, and the trajectory of the sequence is shown in~\figref{fig:trajectory}.

\begin{figure}[t]
	\centering
	\includegraphics[width=1.0\linewidth]{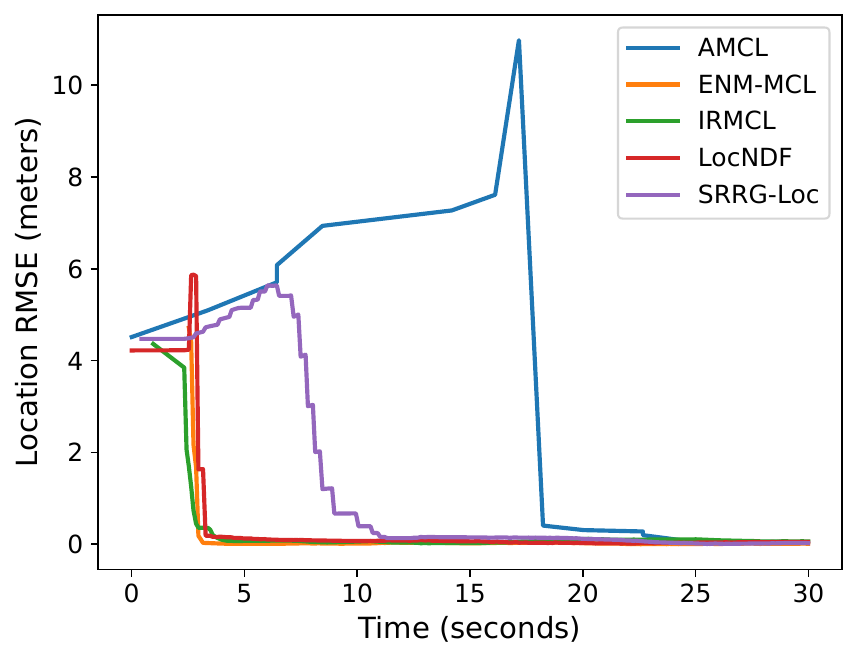}
	\caption{The error curves of location RMSE on the Sequence 5 of the in-house dataset.}
	
	\label{fig:converge_curve}
	\vspace{-0.6cm}
\end{figure}

The location RMSE error curves of all methods are shown in \figref{fig:converge_curve}. 
The occupancy grid map-based methods converge slowly in the sequence because of many similar offices in the environment and the occupancy grid map does not include enough geometric details to quickly distinguish similar rooms. 
In contrast, the implicit representation-based methods quickly converge and our ENM-MCL maintains a fast convergence time of only 3.6 seconds.

In summary, our evaluation suggests that our method is highly efficient and robust, which is suitable for efficient global localization. 
At the same time, our method maintains high accuracy and reliability.
Thus, we supported all our claims with this experimental evaluation.

\begin{table}[t]
	\centering
	\renewcommand\arraystretch{1.2}
		\begin{tabular}{c|ccc}
			\toprule
			\multirow{3}{*}{Method} & \multicolumn{3}{c}{Localization Speed [FPS]} \\
			& Initialization & Pose Tracking & Average Speed \\
			& (\textbf{\#Particles})	& (\textbf{\#Particles}) & on Sequence 1 \\
			\midrule
			IRMCL                   & 1.2 Hz (100,000)      & 27.0 Hz (5,000)     & 9.1 Hz    \\
			LocNDF                  & 0.5 Hz (100,000)      & 2.8 Hz (10,000)    & 2.1 Hz       \\
			ENM-MCL        & \textbf{4.0 Hz (80,000)} & \textbf{250.0 Hz (1,000)} & \textbf{180.2 Hz}     \\
			\bottomrule
		\end{tabular}
	\caption{Runtime comparison of different methods. We report the frame rate for the initialization and pose tracking phase of MCL under the default parameters of baselines. We also report average frame rate of both phases over the complete sequence.
	}
	\vspace{-0.2cm}
	\label{tab:speed}
\end{table}

\section{Conclusion}
\label{sec:conclusion}

In this paper, we presented a novel approach to robot localization using an efficient and effective implicit neural representation.
Our map representation is capable of learning both, the non-projective signed distance fields and direction-aware projective distance fields from 2D LiDAR data, which stores both positional and directional geometric features of the environment. 
Our method exploits a learnable dense feature grid combined with a light-weight neural network as the map representation model.
This allows us to successfully integrate the implicit representation into a Monte Carlo localization framework to improve the accuracy and efficiency for implicit representation-based MCL.
We evaluated our approach on a public dataset and provided comparisons to other existing methods and supported
all claims made in this paper. 
The experiments suggest that by leveraging neural representations, we can not only improve the quality of map, but also reduce the computation costs of MCL to perform real-time pose tracking and enable efficient global localization with a large number of particles.



\bibliographystyle{plain_abbrv}


\bibliography{glorified,new}

\IfFileExists{certificate.tex}{




\onecolumn

~\bigskip\bigskip\bigskip\bigskip 

	\section*{ \LARGE{Certificate of Reproducibility} }\vspace*{1cm}\Large{ 

	The authors of this publication declare that:}

	\vspace{5pt} 

	\begin{enumerate} 

		\setlength{\itemsep}{10pt} 

        \item The software related to this publication is distributed in the hope that it will be useful, support open research, and simplify the reproducability of the results but it comes without any warranty and without even the implied warranty of merchantability or fitness for a particular purpose.
  \item \textit{Haofei Kuang} primarily developed the implementation related to this paper. This was done on  Ubuntu 22.04.

	\item \textit{Yue Pan} verified that the code can be executed on a machine that follows the software specification given in the Git repository available at: \\ 
\begin{center} 

  \url{https://github.com/PRBonn/enm-mcl} \end{center}

 \item \textit{Yue Pan} verified that the experimental results presented in this publication can be reproduced using the implementation used at submission, which is labeled with a tag in the Git repository and can be retrieved using the command:\\ 


\begin{center} 

  \verb|git checkout icra25-release| 

\end{center} 

\end{enumerate} 

\twocolumn

}{}
\end{document}